%% file: root.tex
\documentclass[letterpaper, 10 pt, conference]{ieeeconf}  

\IEEEoverridecommandlockouts                              

\usepackage{graphicx} 
\usepackage{amsmath} 

\usepackage{amsthm}
\usepackage{listings}
\theoremstyle{definition}

\usepackage{amssymb}
\usepackage{pifont}
\usepackage{booktabs}
\usepackage{csvsimple}

\usepackage{caption}
\usepackage{subcaption}
\let\labelindent\relax
\usepackage{enumitem}
\usepackage{bm}
\usepackage{multirow}

\usepackage{algorithm}
\usepackage[noend]{algpseudocode}
\algnewcommand{\LineComment}[1]{\State \(\triangleright\) #1}
\usepackage[colorlinks=true,linkcolor=blue]{hyperref}
\usepackage[font=scriptsize,labelfont=bf]{caption}
\usepackage[dvipsnames]{xcolor}

\usepackage{caption}
\usepackage{subcaption}

\usepackage{makecell}

\usepackage{cite}

\usepackage{adjustbox}

\title{\LARGE \bf
A Causal Approach to Predicting and Improving Human Perceptions of Social Navigation Robots}

\author{Maximilian Diehl$^{1}$, Nathan Tsoi$^{2}$, Gustavo Chavez$^{3}$, Karinne Ramirez-Amaro$^{1}$ and Marynel Vázquez$^{3}$
\thanks{$^{1}$Maximilian Diehl and Karinne Ramirez-Amaro. Faculty of Electrical Engineering, Chalmers University of Technology, SE-412 96 Gothenburg, Sweden.         {\tt\small \{diehlm, karinne\}@chalmers.se}}%
\thanks{$^{2}$Nathan Tsoi. The University of Texas at Austin, Austin, Texas, USA.
    {\tt\small nathan.tsoi@utexas.edu}}%
\thanks{$^{3}$Gustavo Chavez and Marynel Vázquez. Yale University, New Haven, Connecticut, USA.
    {\tt\small \{gustavo.chavez, marynel.vazquez\}@yale.edu}}
}

\begin{document}

\maketitle
\thispagestyle{empty}
\pagestyle{empty}

\begin{abstract}
As mobile robots are increasingly deployed in human environments, enabling them to predict how people perceive them is critical for socially adaptable navigation. Predicting perceptions is challenging for two main reasons: (1) HRI prediction models must learn from limited data, and (2) the obtained models must be interpretable to enable safe and effective interactions. Interpretability is particularly important when a robot is perceived as incompetent (e.g., when the robot suddenly stops or rotates away from the goal), as it allows the robot to explain its reasoning and identify controllable factors to improve performance, requiring causal rather than associative reasoning. To address these challenges, we propose a Causal Bayesian Network designed to predict how people perceive a mobile robot’s competence and how they interpret its intent during navigation. Additionally, we introduce a novel method to improve perceived robot competence employing a combinatorial search, guided by the proposed causal model, to identify better navigation behaviors. Our method enhances interpretability and generates counterfactual robot motions while achieving comparable or superior predictive performance to state-of-the-art methods, reaching an F1-score of $0.78$ and $0.75$ for competence and intention on a binary scale. To further assess our method's ability to improve the perceived robot competence, we conducted an online evaluation in which users rated robot behaviors on a 5-point Likert scale.
Our method statistically significantly increased the perceived competence of low-competent robot behavior by 83\%.
\end{abstract}

\input{main.tex}

\bibliography{references}
\bibliographystyle{IEEEtran}

\end{document}

%% file: main.tex
\section{Introduction}
Robots are becoming more capable, bringing us closer to a future where they will be commonplace in people's lives; however, robots still lack Theory of Mind capabilities~\cite{mao2024review} that facilitate user adoption of the technology. Specifically, it is crucial for robots to model how their performance is perceived by their users during real-world deployments. For example, consider a mobile robot that guides museum visitors to exhibits~\cite{sasaki2017long}. When the robot is perceived as easy to understand, competent, and predictable, its likelihood of adoption increases~\cite{zhang2025predicting}.

Traditionally, human perceptions of robots are measured by surveying users (e.g., with the Robot Social Attribute Scale~\cite{carpinella2017robotic} or the Perceived Social Intelligence Scale~\cite{barchard2020measuring}), but conducting 
surveys requires interrupting the flow of interactions and 
is expensive and time-consuming~\cite{tsoiinfluence}. As an example, the SEAN Together dataset~\cite{zhang2025predicting} with $2,964$ interaction samples between a mobile robot and a human in a Virtual Reality (VR) environment took months to collect. This number of samples is considered large in Human-Robot Interaction (HRI), yet, small for machine learning standards. These challenges motivated recent work on building more scalable, data-driven models that predict human perceptions of robots~\cite{candon2023nonverbal,zhang2023self,zhang2025predicting}.
\begin{figure}[tbp]
\centering\includegraphics[width=0.4\textwidth]{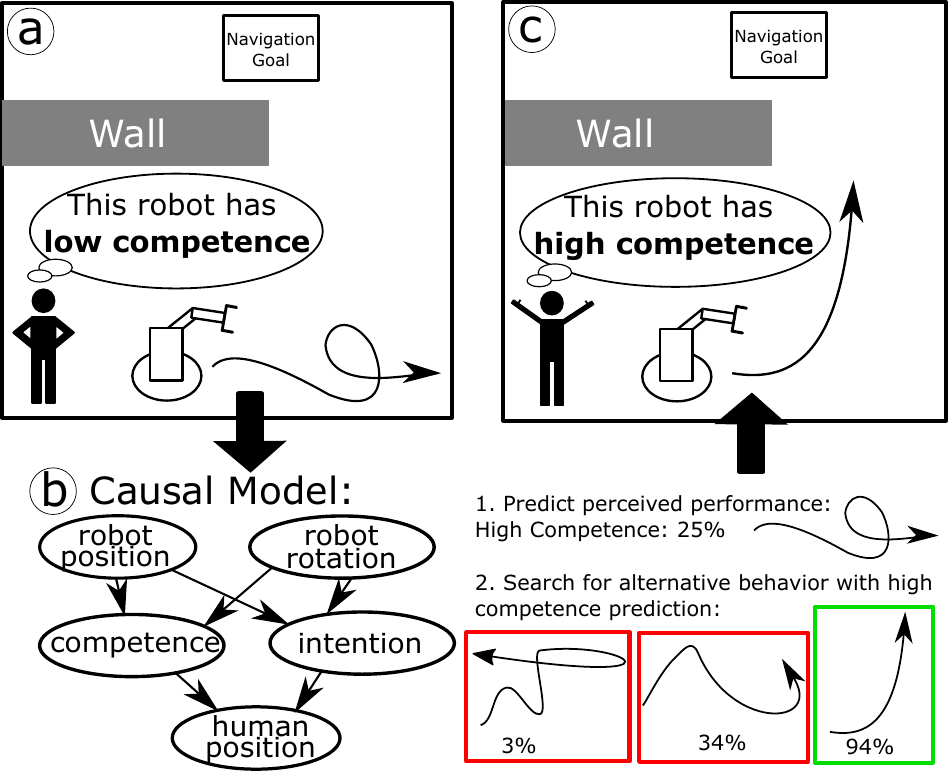}
  \caption{Example of low-competence robot behavior~(a). Our causal model (b) predicts  perceived competence by analyzing environmental cues such as the robot's trajectory. When low competence is predicted, the model identifies the minimal environment change (e.g., robot behavior) that is expected to lead to higher competence~(c).}
\label{fig:fig1}
\vspace{-7mm}
\end{figure}
Unfortunately, existing supervised learning approaches used to predict human perceptions of robots~\cite{candon2023nonverbal,zhang2023self,zhang2025predicting} only model associations among observed features of interactions, making them prone to spurious correlations in the data~\cite{bookOfWhy}. For example, an associative model might suggest that high competence ratings are correlated with a human standing close to the robot, but moving the person closer may not actually improve their perceptions of the robot.

We therefore propose a causal model to predict how humans perceive robots in social navigation scenarios. Causal models are capable of incorporating domain knowledge in their structure, which can reduce the amount of data needed for learning. Moreover, causal models are interpretable and can be used to explain the robot's understanding of the interaction. This allows identifying and reasoning about the causes of a potential failure, as demonstrated in recent studies~\cite{diehl22, diehl23, diehl24}. Our contributions are three-fold: \\
\textbf{1. An interpretable Causal Bayesian Network (CBN) for predicting human perceptions of robot navigation:} Our CBN maps a sequence of observations, such as the robot's trajectory, to a person’s reported impression of the robot’s performance along two dimensions: (i) perceived competence and (ii) understandability of the robot’s intention. To our knowledge, this is the first time causal modeling has been used to predict human perceptions of robots. \\
\textbf{2. A novel method for generating counterfactual robot trajectories:} When the CBN predicts low perceived competence (Figure~\ref{fig:fig1}a), our method generates a counterfactual trajectory via combinatorial search to identify the closest trajectory expected to lead to high perceived competence (Figure~\ref{fig:fig1}b). The robot can then execute this trajectory to preemptively avoid low-competence behavior (Figure~\ref{fig:fig1}c). To accommodate the time-series nature of the data (e.g. SEAN Together dataset), we extend prior work on causal models~\cite{diehl22, diehl23} by integrating trajectory clustering. We also refine the search process to exclude infrequently observed, situation-specific behaviors via thresholding. \\
\textbf{3. Validation of our method's ability to generate alternative robot behaviors that improve perceived robot performance through an online user study}. The study shows that the generated counterfactual behaviors can significantly enhance the perceived competence of a mobile robot.

\section{Related Work}
\subsection{Causality in HRI}
Prior work on causal discovery in robotics has largely focused on manipulation tasks. For example, causal models have been used for tool-affordance learning~\cite{Brawer20IROS} and cube-stacking, where physics simulations were integrated into Causal Bayesian Networks for next-best-action reasoning~\cite{cannizzaro2024causalbayesiannetworkprobabilistic}. In contrast, causal modeling in the HRI domain remains underexplored. Some work has applied causal time series analysis to model human and robot motion based on variables such as agent-goal distance, angle, and velocity~\cite{castri22}, while others investigated improving a robot’s causal understanding by allowing it to ask humans causal questions~\cite{edstrom23}. Our work is uniquely focused on the challenging task of predicting how humans perceive robot navigation. Furthermore, we go beyond modeling motion outcomes by using causal models to generate counterfactual robot behaviors, which can help prevent low perceived competence.

\subsection{Failure Prediction and Prevention in Robotics}
Our work relates to failure prediction and prevention~\cite{honig2018understanding}, as low perceived competence and intention in robot behaviors can be seen as interaction failures that we aim to prevent.
\subsubsection{Predicting perceived agent performance}
Recent work utilized machine learning models such as Support Vector Machines, Random Forests, K-Nearest Neighbors, and Multi-Layer Perceptrons to predict a robot's perceived performance~\cite{zhang2023self, zhang2025predicting}, user preferences for an agent's helping behavior based on nonverbal cues~\cite{candon2023nonverbal}, or models that leveraged social responses to robot errors for timely error detection in human-robot interactions~\cite{stiber2022modeling}. However, such models have limited interpretability due to their black-box nature. Moreover, these methods model correlations rather than causal relationships, which can lead to incorrect predictions when used to intervene in the environment to improve human perceptions of robots~\cite{bookOfWhy}. We address this limitation by proposing an interpretable Causal Bayesian Network (CBN) for predicting how humans perceive a robot’s competence and intention.

\subsubsection{Failures in Robot Manipulation}
FINO-Net~\cite{inceoglu24} proposed a multimodal sensor fusion network to detect and classify failures during manipulation tasks using raw sensory data. REFLECT~\cite{liu2023reflect} used multi-sensory data and a large language model (LLM) to generate failure explanations and create correction plans.  
However, these methods did not consider human perception of robot performance, which are Theory of Mind constructs~\cite{mao2024review}.
Closely related to our contrastive trajectory-generation method, Brandão et al. \cite{brandao21} explained motion-planning failures using contrastive, initialization-based, and trajectory-contrastive techniques. However, their approach lacks a causal model and cannot address perceptions of robot performance in HRI, as it focuses solely on reachability. In contrast, our approach explicitly models human perceptions of robot performance by predicting perceived competence and intention in social navigation scenarios.
\subsubsection{Failures in Navigation}
Robot motion planning is classically defined as finding a collision-free path~\cite{mohanan18}. Recent approaches extend this definition with predicted human motion to ensure safe navigation~\cite{Heuer23, Schaefer21}. As robots increasingly operate in human environments, it becomes important to account for how humans perceive them~\cite{tsoiinfluence}. We address this by generating robot trajectories based on perceived competence and intention. Related work on legible motion~\cite{Dragan15} also incorporates human perception, but  relies on hand-crafted cost functions
~\cite{amirian2024legibot} or black-box learning models
~\cite{wallkotter2022slot}. In contrast, our CBN  learns trajectory classifications directly from data, remaining interpretable, and predicting perceptions along two key dimensions for navigation (competence and intention).


\section{Navigation Task}
\label{sec:task}
We study the Robot-Following task from the SEAN Together dataset~\cite{zhang2025predicting}, in which a robot guides a VR-controlled human through a warehouse environment with obstacles and other moving agents.
The dataset includes observations of nearby autonomously controlled \textbf{agents} (within $7.2$-m, typically considered the robot’s public space~\cite{jensen18}) and the \textbf{human follower}, whose states are represented by relative positions and orientations 
($x,y,\theta$)
with respect to the robot. In addition, it contains the \textbf{goal position}, specified by relative 
$(x,y)$ coordinates, and a local \textbf{occupancy map} (a $7.2$-m × $7.2$-m crop centered on the robot) encoding nearby static obstacles via a ResNet-18~\cite{he2016deep} embedding. Additionally, the human-follower was asked to rate \textit{how \textbf{competent} the robot was at navigating} and \textit{how clear the robot’s \textbf{intentions} were during navigation} on a 5-point Likert scale.\footnote{The dataset also contains ratings for how \textit{surprising the robot's navigation behavior was}, which is not considered in our work, as surprise can be interpreted positively or negatively, depending on the context.} To collect these ratings, the navigation was paused at random intervals, and a rating screen was displayed within the VR environment. In our work, we want to modify the robot's behavior when the person perceives the robot as performing poorly (e.g., low competence).
We therefore transformed the 5-point format to a binary rating (1–3 = low (0), 4–5 = high (1)). SEAN Together~\cite{zhang2025predicting} has 2,964 samples from 60 people, $\mathcal{D}={(o_{1:T}, y_i)}$. Each sample has 8-second time-series data over all previously described observation variables ($o_{1:T}$) and a single performance rating gathered at the end of that sequence ($y_i$). 
We propose a CBN to infer $y_i$ given $o_{1:T}$. 

\section{Method}
\subsection{Causal Bayesian Networks}
Formally, CBNs are defined as \textit{Directed Acyclic Graphs} (DAG)  $\mathcal{G} = (\boldsymbol{X}, A)$, where the nodes $\boldsymbol{X} = \{ X_1, X_2, $ $..., X_N \}$ are a set of $N$ random variables $X_i \subseteq\  \boldsymbol{X}$, and $A$ is the set of arcs~\cite{bnlearn} that describe the causal connections between the variables. We model the Robot-Following task as a CBN, where $\boldsymbol{X}$ represents the task variables as specified in Section~\ref{sec:graph}. Based on the dependency structure of the DAG and the \textit{Markov property}, the \textit{joint probability distribution} of a causal BN can be factorized into a set of \textit{local probability distributions}, where each $X_i$ only depends on its direct parents $\text{Pa}_{X_i}$:
\vspace{-0.5em}
\begin{equation}\label{eq:bn} 
    P(X_1, X_2, ..., X_N) = \prod_{i=1}^{n}P(X_i|\text{Pa}_{X_i})
\end{equation}
\vspace{-0.75em}

Traditionally, CBNs are obtained in two steps: \textit{Structure Learning} and \textit{Parameter Estimation}. 

\subsubsection{Structure Learning}
With sufficient data, the causal structure of a CBN $\mathcal{G} = (\boldsymbol{X}, A)$ can be learned using algorithms such as Grow–Shrink~\cite{gs} or PC~\cite{pcstable}, constraint-based methods that utilize statistical tests to identify conditional independence relations. However, learning plausible causal relationships is challenging~\cite{Sharma21}, particularly with limited data, as in our HRI setting. In such cases, the causal graph is typically specified by a domain expert. The advantage of the separation in structure learning and parameter learning is that the expert only needs to provide high-level causal relationships rather than probability distributions.

\subsubsection{Parameter Estimation}
Once the structure $\mathcal{G}$ is defined, the local conditional distributions $P(X_i|\text{Pa}_{X_i})$ (Eq.~\ref{eq:bn}) are typically estimated by Maximum Likelihood Estimation (MLE) or Bayesian estimation~\cite{Ji15}. In our case, all variables $X_i$ are multinomial, thus each local distribution can be represented as a table of parameters~$\boldsymbol{\theta}$. Each entry $\theta_{x|\boldsymbol{u}} \in \boldsymbol{\theta}$ 
gives the probability $\theta_{x|\boldsymbol{u}} = P(X_i=x|\text{Pa}_{X_i}=\boldsymbol{u})$ of $X_i$ taking value $x$ for a particular parent configuration $\boldsymbol{u}$. The size of this table grows with the number of parents and their number of discretization intervals.

\subsubsection{Preventing Robot Failures}
Our method for preventing low perceived competence builds on prior work that uses CBNs to prevent robot task-execution failures~\cite{diehl22, diehl23, diehl24}. The core idea is to use the CBN to predict the probability of a desirable outcome, e.g., high perceived competence, based on the current values of the variables that causally impact the competence. If this predicted probability falls below a threshold $\epsilon$, we search for alternative parent-variable values that would yield high perceived competence. 
However, the existing approaches cannot be directly applied to the SEAN Together dataset for two main reasons. First, the dataset contains time-series variables, such as the robot’s trajectory and orientation, whereas prior methods assume all variables are single-valued and discretized. Second, unlike the previous work, which relied on large-scale simulated datasets to learn both the causal structure and the associated conditional distributions, the SEAN Together dataset contains only a limited number of real-world human-robot interactions. This makes it substantially more difficult to infer a reliable causal structure and to estimate the conditional distributions directly from data. In the following subsections, we present our proposed method and discuss how it addresses these challenges.

\subsection{Proposed CBN for the Robot-Following Task}
\label{sec:graph}

\subsubsection{Data Preprocessing} 
Our modeling process had three objectives: (1) limit the number of parent variables per node to keep parameter estimation feasible, (2) ensure each node has a clear semantic meaning for interpretability, and (3) make nodes, particularly those influencing competence and intention, actionable by the robot. To achieve this, we manually applied three key modifications to the existing variables. First, all distance variables originally measured in $(x, y)$ coordinates were combined into a single $L^2$ norm, applied to both robot-goal and human-robot distances. 
Second, we converted all distances, originally measured as absolute values, into relative changes with respect to the first measurement in each 8-second time series. In this formulation, each distance time-series trajectory begins at 0 (representing the current location) and indicates how the distance changes over the 8-second interval. This ensures all trajectories are physically executable. For example, if our method were to recommend an alternative robot trajectory with a different initial distance to the goal, executing such a trajectory would be infeasible without ``teleporting'' the robot to a new initial state. Third, variables related to autonomous pedestrians and map information were removed. While this variable selection remains manual and task-dependent, it reduces model complexity and improves interpretability. The final set of variables is shown in Tbl.~\ref{tab:variables}.

\input{tables/tbl1}

\subsubsection{Generalized Discretization through time-series clustering}

A major contribution of our work is increasing the flexibility of existing models to incorporate both time-varying and static variables. To achieve this, we propose a generalized discretization process (Alg.~\ref{alg:timeseries}) that can discretize all variables defined in Tbl.~\ref{tab:variables}. 

The inputs to Alg.~\ref{alg:timeseries} are the training dataset $\mathcal{D}$ (consisting of $K$ IID samples $\boldsymbol{\xi}$, each a fully observed instance of all network variables $\boldsymbol{X}$), and $\Lambda$, which specifies the number of discretization intervals per variable. The output is a list of discretization intervals $X_{\text{int}}$ for each variable. 

For time-series variables $\mathcal{D}_i$ ($\mathcal{D}_i$ denotes all observed values of a specific variable $X_i$) we apply K-means clustering~\cite{scikit-learn} (Alg.~\ref{alg:timeseries}, Line 6), treating each sequence as a vector of data points and grouping similar series by Euclidean distance into $\Lambda_i$ clusters $\{ I_1, \dots, I_{\Lambda_i} \}$. The cluster centroids $\{ c_1, \dots, c_{\Lambda_i} \}$ then define the discretization intervals, representing average patterns. For continuous single-valued variables, we perform quantile discretization~\cite{bnlearn} (Alg.~\ref{alg:timeseries}, Line 8), dividing the data into $\Lambda_i$ intervals with equal numbers of data points. Categorical variables are discretized by directly assigning the unique values as intervals (Alg.~\ref{alg:timeseries} Line 10). For inference at runtime, each time-series variable (trajectory over the previous 8 seconds) is assigned to the nearest cluster centroid using Euclidean distance.

\begin{algorithm}[tbhp]
	\caption{\label{alg:timeseries}\textsc{GetVariableIntervals} function}
	\begin{algorithmic}[1]
		\Function{GetVariableIntervals}{$\mathcal{D}$, $\Lambda$}
		\State $X_{\text{int}} \leftarrow []$
		\ForAll{$i \in |\boldsymbol{X}|$}
		\If{$\mathcal{D}_i$ \textbf{is not categorical}}
		\If{$\mathcal{D}_i$ \textbf{is time-series}} 
		\State $X_{\text{int}_i} \leftarrow$ 
		\textsc{Add}$(X_{\text{int}_i}, \textsc{Cluster}(\mathcal{D}_i, \Lambda_i))$
		\Else
		\State $X_{\text{int}_i} \leftarrow$ 
		\textsc{Add}$(X_{\text{int}_i}, \textsc{Discretize}(\mathcal{D}_i, \Lambda_i))$
		\EndIf
		\Else 
		\State $X_{\text{int}_i} \leftarrow$
		\textsc{Add}$(X_{\text{int}_i}, \textit{Val}(\mathcal{D}_i))$
		\EndIf
		\EndFor
		\State \textbf{return} $X_{\text{int}}$
		\EndFunction
	\end{algorithmic}
\end{algorithm}

\subsubsection{Proposed CBN}
\begin{figure}[tbhp]
\centering
\vspace{-0.7em}
  \includegraphics[width=0.38\textwidth]{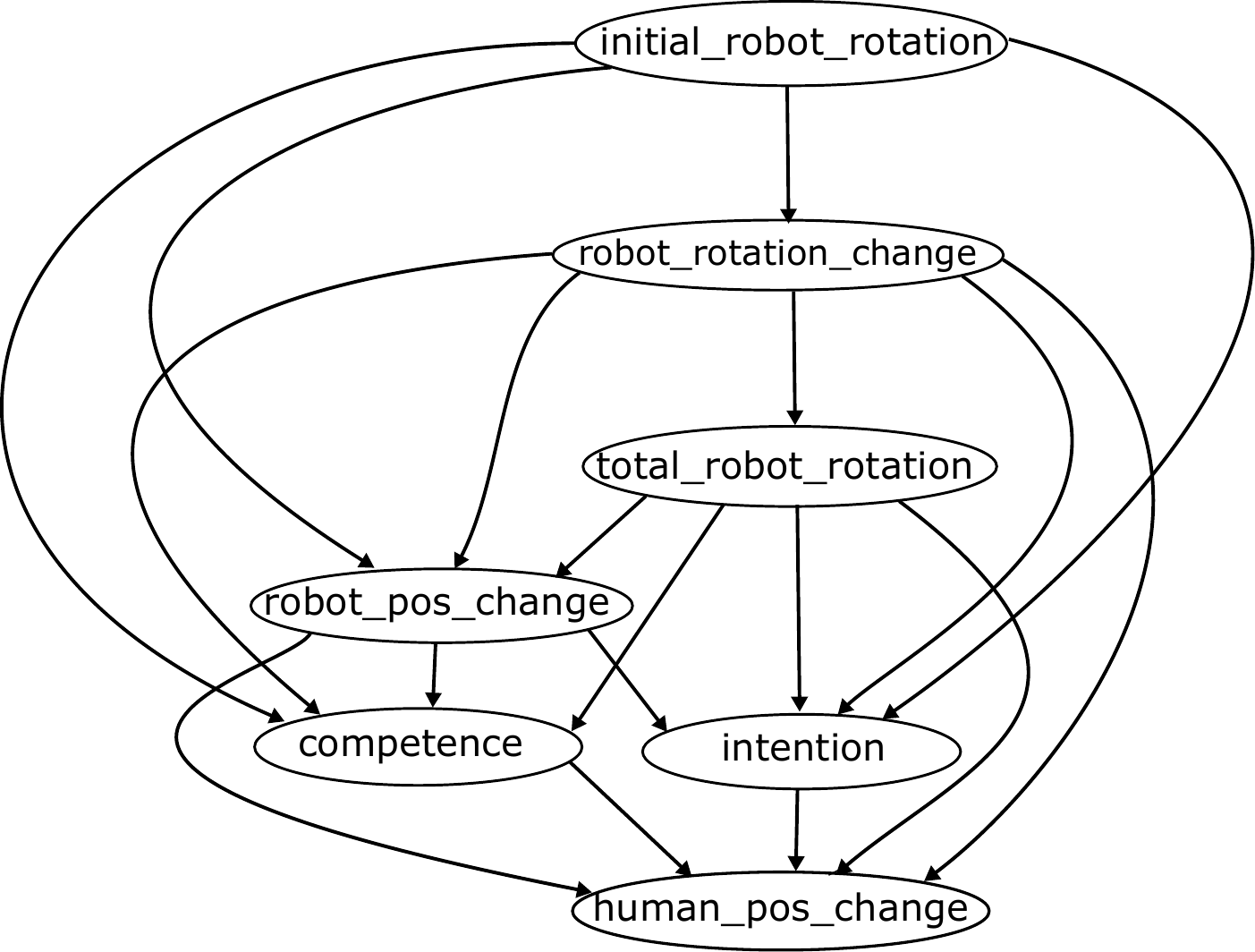}
  \caption{Proposed CBN graph for the Robot-Following task.}
\label{fig:graph}
\end{figure}

Our proposed CBN structure $\mathcal{G}$ for the Robot-Following task is visualized in Fig.~\ref{fig:graph}. Its variables $\boldsymbol{X}$ 
are defined as in Tbl.~\ref{tab:variables} and discretized via Alg.~\ref{alg:timeseries}.

Our model includes three different aspects of the robot's rotation:
The key variable describing the rotational trajectory is \texttt{robot\_rotation\_change}. However, this alone is insufficient for reliably estimating competence or intention, as it measures rotation relative to the start of an 8-second interval and does not capture the initial misalignment toward the goal as measured through \texttt{initial\_robot\_rotation}. For instance, a robot already facing the goal should maintain orientation, whereas a misaligned robot should rotate toward it. We also observed that our clustering struggled to distinguish trajectories that maintain orientation from those where the robot rotates around its own axis, since both have similar net rotation toward the goal. We therefore introduced \texttt{total\_robot\_rotation}, capturing the cumulative rotation over the interval. 

Another key variable affecting perceived competence and intention is \texttt{robot\_pos\_change}, which represents the robot’s movement relative to the goal. As our robot is not omnidirectional, its movement is partially constrained by rotation: if the robot spins in place, it cannot advance, whereas proper orientation toward the goal allows movement toward the goal. 

Human movement (\texttt{human\_pos\_change}) is modeled as a direct consequence of perceived competence, perceived intention, and the robot’s motion given that the human  follows the robot in the Robot-Following task~\cite{zhang2025predicting}.  

\subsection{Proposed method to address low perceived performance}
We propose a new method for failure explanation and prevention (Alg.~\ref{alg:explanation}) that extends the approach of~\cite{diehl22, diehl23} to generate counterfactual robot behaviors. When the current behavior $x_{\text{current}_\text{int}}$ is predicted to lead to low competence, Alg.~\ref{alg:explanation} identifies alternative variable parameterizations that are expected to improve human perception given the CBN.

\setlength{\textfloatsep}{2pt}
\begin{algorithm}[tbhp]
  \caption{\label{alg:explanation}Counterfactual Robot Behavior Generation}
  \hspace*{\algorithmicindent} \textbf{Input:} discretized failure variable parameterization $x_{\text{current}_\text{int}}$, graphical model $\mathcal{G}$, structural equations $P(X_i|\text{Pa}_{X_i})$, discretization intervals of all model variables $X_{\text{int}}$, success threshold $\epsilon$, goal parametrizations $X_{\text{goal}}$ \\
  \hspace*{\algorithmicindent} \textbf{Output:} solution variable parameterization $x_{\text{solution}_{\text{int}}}$, solution success probability prediction $p_{\text{solution}}$
  \begin{algorithmic}[1]
  \State $P \leftarrow \textsc{generateTransitionMatrix}(X_{\text{int}})$ 
  \State $q \leftarrow [x_{\text{current}_{\text{int}}]}$
  \State $v \leftarrow []$
  \While{$q \neq \emptyset$} 
   \State $node \leftarrow \textsc{Pop}(q)$ 
   \State $v \leftarrow \textsc{Append}(v, node)$
   \ForAll{$child$ reachable from $node$ in $P$}
   \If{$child \not\in q,v 
  \quad \textbf{and} \quad |\text{Pa}_{\text{child}}| > m$}
   \State $p_{\text{solution}} = P(\text{competence}=1|\text{Pa}_{child})$
   \If{$p_{\text{solution}} > \epsilon$}
   \State $x_{\text{solution}_{\text{int}}} \leftarrow child$
   \State $\textsc{return}(p_{\text{solution}}, x_{\text{solution}_{\text{int}}})$
   \EndIf
   \State$ q \leftarrow \textsc{Append}(q, child)$
   \EndIf
   \EndFor
   \EndWhile
  \end{algorithmic}
\end{algorithm}

Line 1 of Alg.~\ref{alg:explanation}  initializes a matrix that defines all valid transitions for each possible parameterization $\boldsymbol{u}$ of the parent variables of the outcome (e.g., competence or intention). The matrix guides the search by allowing only minimal changes, in line with the Occam's Razor principle to find the smallest necessary change. For example, with two parent variables $X$ and $Y$, each with four intervals ($X = \{x_1, x_2, x_3, x_4\}$ and $Y = \{y_1, y_2, y_3, y_4\}$), a valid transitions from $node = (X = x_1, Y = y_4)$ would be to $child_1 = (X = x_2, Y = y_4)$ or $child_2 = (X = x_1, Y = y_3)$. These adjust only one variable by one neighboring interval. In contrast, transitions that change multiple variables simultaneously or skip intervals (e.g., 
$x_1$ directly to $x_3$) are disallowed.

Specifically, in the context of our robot-follower task, a variable parameterization refers to a specific combination of the initial robot rotation interval, total robot rotation interval, rotation cluster, and position cluster ($\text{Pa}_{\text{competence}}$). In our implementation, we prioritize changes to the \texttt{robot\_pos\_change}, \texttt{robot\_rotation\_change}, and \texttt{total\_robot\_rotation} variables before modifying \texttt{initial\_robot\_rotation}. This means our algorithm first searches for possible interventions on robot behavior within the subset of variables that maintain the same initial rotational difference. Only if no successful alternatives are found within this subset does the algorithm consider other initial rotations.
To identify the optimal counterfactual behavior, we adapt a Breadth-First Search (BFS) procedure that explores variable parameterizations in search of one that fulfills the target competence or intention criteria. The search terminates upon finding a parameterization $child$ that satisfies the success condition $P(\text{competence} = 1 | \text{Pa}_{child})> \epsilon$, where $\epsilon$ represents a threshold indicating a sufficiently high likelihood of achieving the desired competence. This is done in Lines 4-13 of Alg.~\ref{alg:explanation}.
Critically, 
to more reliably deal with small datasets, Line 9 introduces a modification where a parameterization is only considered if it has been observed a sufficient number of times in the training dataset $\mathcal{D}$. Specifically, the search condition is entered only if the count of observations for the parent configuration $\text{Pa}_{child}$, denoted $|\text{Pa}_{child}|$ exceeds a threshold $m$. This requirement ensures that the algorithm relies only on parameterizations with sufficient support in the data, thereby filtering out infrequent or anomalous behaviors that may not generalize well. In our experiments, we set $m=5$, which was heuristically determined to filter out infrequent behaviors that are unlikely to be effective as general solutions for preventing low perceived competence, and set $\epsilon=0.9$ to ensure high success probability among the counterfactual solution.

The final output of Alg.~\ref{alg:explanation} is a new variable parameterization, $x_{\text{solution}}$, for which the model predicts a high rating (e.g., positive competence). This solution provides alternative robot position and trajectory behaviors in the form of cluster centroids and intervals. By implementing these behaviors on the robot, 
it is expected to be perceived more positively.
\section{Clustering and Prediction Performance}
In this section, we first describe the clusters obtained from the time-series data provided by SEAN Together~\cite{zhang2025predicting} and then compare the prediction performance of our Causal Model with that of a Random Forest classifier. The Random Forest was previously identified as the best-performing baseline model among several alternatives for this data~\cite{zhang2025predicting}, including a Graph Neural Network and a Multi-Layer Perceptron.

We evaluate all models using F1-Score, Accuracy, Precision, and Recall. These metrics were computed using Leave-One-Out Cross-Validation (LOOCV). In each run, we held out all samples from one participant for testing and trained the model on the remaining participants. The final performance values are the macro-averaged results across all participants. For the Causal Model, we performed an additional hyperparameter search over the number of intervals $\Lambda$ using nested LOOCV. With 60 participants, we exhaustively iterated over all splits of 58 for training, 1 for validation, and 1 for testing. The best hyperparameters were selected based on the macro-averaged F1-Score. The final reported results were obtained by running LOOCV once more using these selected hyperparameters.

\begin{figure}[tbhp]
\centering
  \includegraphics[width=0.47\textwidth]{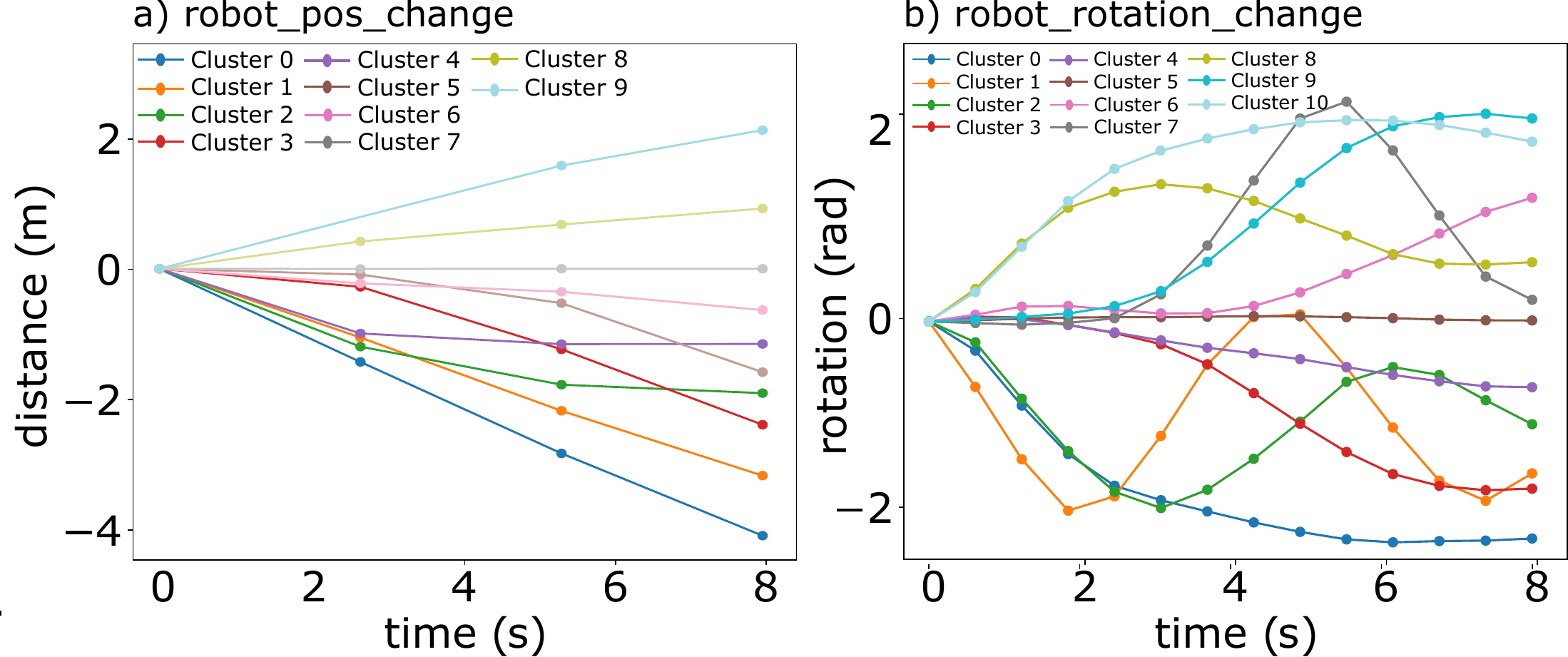}
  \caption{Cluster means: robot\_pos\_change and robot\_rotation\_change.}
\label{fig:cluster_means}
\end{figure}

The hyperparameter search yielded four intervals for both \texttt{initial\_robot\_rotation} and \texttt{total\_robot\_rotation}, and ten and eleven clusters for \texttt{robot\_pos\_change} and \texttt{robot\_rotation\_change}, respectively (Fig.~\ref{fig:cluster_means}). 
The position and orientation clusters capture intuitive motion patterns, ranging from increasing (Cluster 9, Fig.~\ref{fig:cluster_means}a), constant (Cluster 7, Fig.~\ref{fig:cluster_means}a), or decreasing distance to the goal (Cluster 0, Fig.~\ref{fig:cluster_means}a) to maintaining orientation (Cluster 5, Fig.~\ref{fig:cluster_means}b), rotating toward (Cluster 0, Fig.~\ref{fig:cluster_means}b) or away from the goal (Cluster 10, Fig.~\ref{fig:cluster_means}b).

In Tbl.~\ref{tab:results}, we compare the prediction performance of our causal model with the best-performant Random Forest (RF) classifier (Nav. + Facial features) from~\cite{zhang2025predicting}. Measured by Accuracy, our model outperformed the baseline by $0.021$ for Competence and $0.029$ for Intention. Measured by F1-Score our method outperformed the baseline by $0.047$ and $0.044$ for Competence and Intention, respectively.

In conclusion, on average, our method outperformed the prior state-of-the-art, black-box machine learning approach. Our causal model has the additional benefit that it is interpretable and simpler, using only a subset of the full feature set from  SEAN Together. Moreover, the model encodes causal information, which is crucial for generating behaviors for improving the robot's perceived performance.

\input{tables/tbl2}

\section{User Study of Counterfactual Behavior Generation}
\label{sec:study}

We conducted an online user study to evaluate the capability of our method
to generate counterfactual trajectories with greater \textit{perceived robot competence}, using the Prolific crowdsourcing platform. Our study was reviewed and approved by our university's ethics review board.

\subsection{Hypotheses}
We hypothesized that our method can improve the perceived competence of robot navigation behavior, specifically: \\
\textbf{H1:} When our causal model predicts the perceived competence \textit{correctly as low}, our approach generates navigation behaviors that are perceived as more competent than the original robot behavior.\\
\textbf{H2:} When our causal model predicts the perceived competence \textit{erroneously low}, our approach still generates navigation behaviors that are perceived as more competent than the original robot behavior.

To test both hypotheses, we implemented two distinct study phases: One phase included 10 scenarios in which our model correctly predicted low robot competence. The other phase studied 10 scenarios where the model incorrectly classified the robot as low competent. We recruited a different set of participants for each study phase.

\subsection{Study Phases and Conditions}
\begin{figure*}[tbhp]
	\centering
	\begin{subfigure}[b]{\textwidth}
		\centering
		\includegraphics[width=0.9\textwidth]{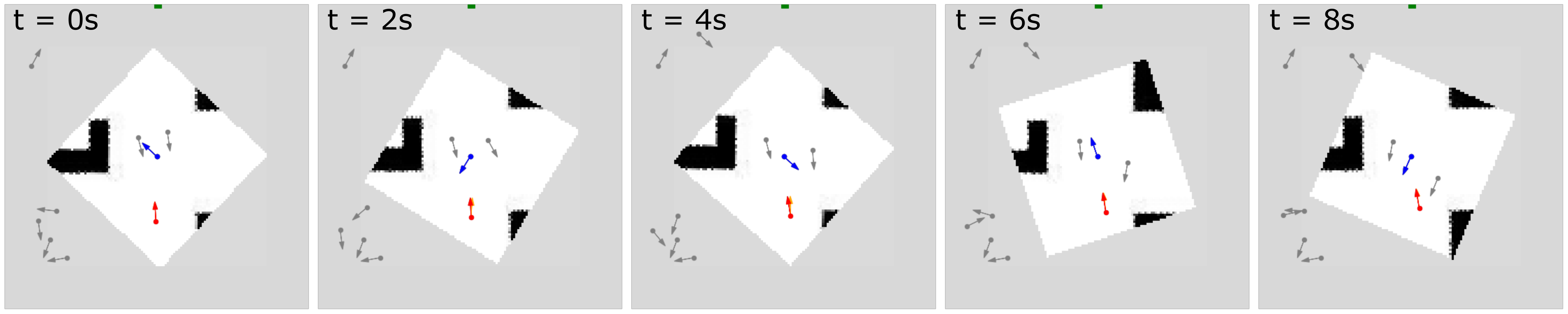}
		\caption{Original robot behavior}
		\label{fig:original}
	\end{subfigure}
	\begin{subfigure}[b]{\textwidth}
		\centering
    \vspace{0.3em}
		\includegraphics[width=0.9\textwidth]{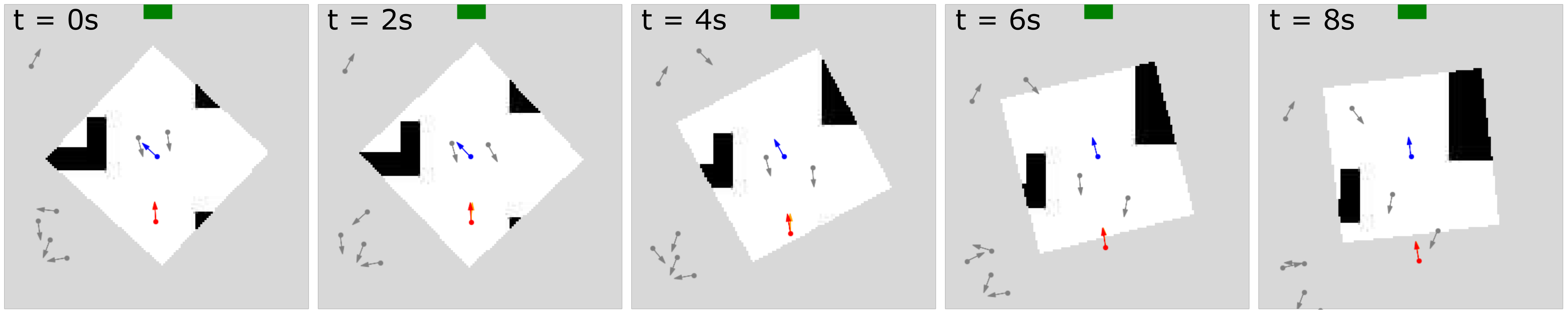}
		\caption{Counterfactual robot behavior}
		\label{fig:counterfactual}
	\end{subfigure}\caption{\label{fig:topdown_navigation}Image series of a navigation-task video for our user study (Sec.~\ref{sec:study}). The blue arrow represents the robot’s position and orientation, the red arrow depicts the follower, and other agents are shown as grey arrows. The goal, at the top of the images, is a green rectangle. Black areas indicate static obstacles, white areas are navigable space, and the surrounding grey region lies outside the 7.2m public space~\cite{jensen18} where the 2D map was recorded. The upper series shows the original robot behavior, classified as low competence by our model, while the lower series depicts the counterfactual behavior generated to address this low-competence trajectory. Images are best viewed in color.}
    \vspace{-1.5em}
\end{figure*}

Both study phases included two conditions: ``Original'', and ``Counterfactual.'' In the Original condition, participants viewed videos directly from the SEAN Together dataset (Fig.~\ref{fig:original}). In the Counterfactual condition, videos were generated by modifying the robot’s behavior to follow the average cluster means of alternative \texttt{robot\_pos\_change} and \texttt{robot\_rotation\_change} clusters as obtained through our proposed method (Fig.~\ref{fig:counterfactual}). Position trajectories (e.g., decreasing the robot's distance towards the goal by $4m$ as shown by Cluster 0 in Fig.~\ref{fig:cluster_means}b) were adjusted along a straight line toward the goal. 
Rotations were down-scaled if smaller than the cluster maximum but not increased; as a result, in some cases, the robot moved toward the goal without fully aligning its orientation. Position and rotation trajectories were synchronized in time but applied independently. To maintain visual consistency with the original videos, the robot remained centered, while the map, pedestrians, and human follower positions were adjusted. We could not predict how the human follower and pedestrians would respond to the counterfactual robot behavior and, thus, kept their movements unchanged.

\subsection{Experimental Procedure}
Participants completed the online study using a standard desktop web browser; mobile devices were not permitted. After providing consent, they completed a brief demographic survey and read study instructions, which explained the SEAN Together dataset~\cite{zhang2025predicting}, the task of rating robot trajectories, the expected duration ($\approx$15 minutes)
and payment (USD \$3.00). Participants were familiarized with the top-down video perspective by viewing an example video.
Each participant viewed 20 videos in total: 10 showing the original robot behavior and 10 showing counterfactual behaviors. These videos were generated from a random sample of 10 scenarios from a 120-scenario test set not included in training. All participants saw the same 10 scenarios (both original and counterfactual), but the order was randomized to reduce ordering effects.
After each video, participants rated how they believed the human follower perceived the robot’s competence on a 5-point Likert scale (from 1 = “very incompetent” to 5 = “very competent”). The supplementary video provides a visual overview of the study.

For each study phase, we recruited 40 participants (80 total) from North America, aged 18 or older, with normal or corrected-to-normal vision, and fluent in English. Their mean age was $37.21 \pm 12.30$ years, with 13 female and 27 male participants. On average, participants reported near-neutral familiarity with robots ($2.94 \pm 1.41$ on a 7-point scale, where 1 = “Not at all familiar” and 7 = “Very familiar”).

\subsection{Results}


We fitted linear mixed-effect models 
for the competence measure with fixed effects for the condition (Original or Counterfactual). A linear mixed-effect model was used due to the hierarchical nature of the data, i.e., Participant ID was nested within Scenario ID and added as a random effect to the model.
This nesting associated the robot's behavior across the two instances of a scenario shown to each participant.

\vspace{0.5em}
\noindent\textbf{(H1) Competence of the Counterfactuals when the CBN model \textit{correctly} predicted the Original robot behavior as low competent.} We found a significant difference in participants' ratings between the original trajectories and our model's proposed counterfactual trajectories when our CBN model made a correct prediction, $F(1, 399)=338.36, p<0.0001$. Overall, participants rated the competence of counterfactual trajectories ($3.59 \pm .06$) higher than the original trajectories ($1.96 \pm 0.06$).
Therefore, on average, counterfactual trajectories improved the perceived competence of the robot by $83\%$.
These results support H1, demonstrating our method's potential to improve robot trajectories and enhance the perceived competence of a robot.

\vspace{0.5em}
\noindent\textbf{(H2) Competence of the Counterfactuals when the CBN model \textit{erroneously} predicted the Original robot behavior as low competent.} We found a significant difference in participants' ratings between the original trajectories and our model's proposed counterfactual trajectories when our CBN model made an incorrect prediction, $F(1, 399)=76.32, p<0.0001$. Overall, participants rated the competence of counterfactual trajectories ($3.42 \pm .07$) higher than the original trajectories ($2.69 \pm 0.07$). Thus, our counterfactual trajectories improved the perceived competence of the robot by $27\%$.
These results support H2, demonstrating our method's potential to improve robot trajectories and enhance the perceived competence of the robot, even when the model makes an incorrect initial prediction regarding the robot's perceived competence.

\subsection{Discussion}
Our results suggested that, across both study phases, the counterfactual trajectories were rated significantly more competent than the original robot behaviors. In terms of absolute scores, the competence ratings for the generated counterfactuals were consistent between the two study phases, with an overall average rating of $3.51$ over the 5-point competence scale. 
Although the original robot trajectories in the second study phase ($2.69 \pm 0.07$) were rated higher than in the first phase ($1.96 \pm 0.06$), their average ratings were below the mid-mark of the 
competence scale. 
This suggests that the proposed causal model is capable of identifying additional cases where a correction of the robot's trajectory may be needed for an average user.

\section{Limitations \& Future Work}

To enable our CBN model to effectively work on small datasets, we excluded environmental factors such as walls and pedestrians from the causal graph. This simplification helped reduce model complexity and prevent overfitting. Even though the simplification lead to breaking the assumption of no unobserved confounding variables~\cite{bookOfWhy}, we observed only minimal negative effects on the robot’s behavior in our study: In one scenario (out of the 20 tested), the counterfactual trajectory guided the robot to rotate and move toward a wall rather than around it, reflecting the model’s general tendency to favor direct, goal-oriented motion over low-competence actions such as rotating away from the goal. 

Another factor, which might have influenced our study results, was the missing ground truth motion of the human follower in the case of  counterfactual robot navigation. We had to re-use the original human trajectories for counterfactual navigation, as  in Fig.~\ref{fig:topdown_navigation}b, where there is an increasing gap between the robot and the follower. This gap could have been interpreted as the robot moving too quickly or not coordinating sufficiently with the human. Future work could  integrate the trajectories generated by the causal model into a motion planner for the robot, which would allow  to incorporate environmental constraints such as obstacles and pedestrian flow during motion generation. This integration could also enable replication of the study in a VR-based environment, where human follower ratings could be collected directly from the participants for counterfactual robot motion, eliminating the need for follower ground-truth trajectories.

While previous work on causal modeling~\cite{diehl22} has learned the structure of causal graphs given sufficiently large datasets (e.g., 20,000 samples), we relied on manual specification of the causal graph in this work (while learning all probability distributions for the CBN). 
To further reduce manual effort when transferring the approach to new HRI domains, we are excited to explore the use of Large Language Models (LLMs) in the future. LLMs  have recently been shown to assist in proposing candidate causal structures~\cite{jiralerspong2024}, and could aid in data-limited regimes, as is often the case in HRI.

Finally, it would be interesting to expand our investigations on improving robot competence to other human perceptions. For example, we did not try to model how surprising the robot behavior was during navigation, because surprising behavior could be both positive or negative. But, in the future, it would be interesting to model this factor as a two-dimensional construct, with  a rating and a valence (positive or negative). Also,  \cite{tsoiinfluence} studied other perceptions important for navigation, including perceived robot intelligence~\cite{barchard2020measuring}, which could be modeled with a CBN in the future.

\section{Conclusion}
We proposed a Causal Bayesian Network for online prediction of human perceptions of a navigation robot's competence and intention. The model outperformed the state-of-the-art supervised correlative model in predicting perceived robot competence and intention while remaining interpretable. We also proposed a new method for generating counterfactual robot behaviors that improved the perceived competence of low-competence robot navigation by 83\%.

%% file: tables/tbl1.tex
\begin{table*}[tbhp]
\centering
\resizebox{\textwidth}{!}{
\begin{tabular}{|p{2.5cm}|p{8cm}|p{5.5cm}|}
\hline
\textbf{Variable Name} & \textbf{Formula} & \textbf{Interpretation \& Type} \\
\hline
robot\_rotation\_change &
$\{\theta^{\text{robot\_goal}}_{0}-\theta^{\text{robot\_goal}}_{0}, \dots, \theta^{\text{robot\_goal}}_{8}-\theta^{\text{robot\_goal}}_{0}\}$ &
\makecell[l]{Is the robot rotating towards/away from the\\ goal? (time-series)} \\
\hline
total\_robot\_rotation &
$\sum_{t=0}^{8} |\theta^{\text{robot\_goal}}_t|$ &
\makecell[l]{Total robot rotation (over 8 second\\ observation) (continuous)} \\
\hline
initial\_robot\_rotation &
$\theta^{\text{robot\_goal}}_0$ &
\makecell[l]{Initial robot-goal angle (continuous)} \\
\hline
robot\_pos\_change &
$\{\text{dist}^{\text{robot\_goal}}_{0}-\text{dist}^{\text{robot\_goal}}_{0}, \dots, \text{dist}^{\text{robot\_goal}}_{8}-\text{dist}^{\text{robot\_goal}}_{0}\}$ &
\makecell[l]{Is the robot moving towards/away from the\\ goal? (time-series)} \\
\hline
competence &
$\{0,1\}_{t=8}$ &
\makecell[l]{Perceived competence at the end of an\\ observation (categorical)} \\
\hline
intention &
$\{0,1\}_{t=8}$ &
\makecell[l]{Perceived intention at the end of an\\ observation (categorical)} \\
\hline
human\_pos\_change &
$\{\text{dist}^{\text{human\_robot}}_{0}-\text{dist}^{\text{human\_robot}}_{0}, \dots, \text{dist}^{\text{human\_robot}}_{8}-\text{dist}^{\text{human\_robot}}_{0}\}$ &
\makecell[l]{Is human moving towards/away from the\\ robot? (time-series)} \\
\hline
\end{tabular}
}
\caption{CBN variables $\boldsymbol{X}$. $\theta^{\text{robot\_goal}}_t$ denotes the angle between robot and goal and $\text{dist}_t$ denotes the Euclidean distance at time $t$.}
\label{tab:variables}
\vspace{-1em}
\end{table*}

%% file: tables/tbl2.tex
\begin{table}[tbhp]
    \centering
    \begin{adjustbox}{max width=\columnwidth}
    \begin{tabular}{|c|c|c|c|c|}
        \hline
        \textbf{Method} & \textbf{F1} & \textbf{Accuracy} & \textbf{Precision} & \textbf{Recall} \\
        \hline
        \hline
        \multicolumn{5}{|c|}{\textbf{Competence}} \\
        \hline
         Causal (ours) & $\mathbf{0.777 \pm 0.09}$ & $\mathbf{0.835 \pm 0.08}$ & $0.811 \pm 0.12$ & $\mathbf{0.768 \pm  0.14}$\\
         \hline
          RF & $0.73 \pm 0.12$ & $0.814 \pm 0.09$ & $\mathbf{0.816 \pm 0.13}$ & $0.686 \pm 0.16$ \\
        \hline
        \hline
        \multicolumn{5}{|c|}{\textbf{Intention}} \\
        \hline
        Causal (ours) & $\mathbf{0.751 \pm 0.1}$ & $\mathbf{0.788 \pm 0.10}$ & $\mathbf{0.823 \pm 0.11}$ & $\mathbf{0.713  \pm 0.15}$\\
        \hline
        RF & $0.707 \pm 0.12$ & $0.759 \pm  0.12$ & $0.817 \pm 0.11$ & $0.654 \pm 0.18$\\
        \hline
    \end{tabular}
    \end{adjustbox}
    \caption{LOOCV evaluation ($\mu \pm \sigma$) on binary F1-score, Accuracy, Precision, and Recall. Our Causal Model is compared against the best-performing Random Forest baseline from related work (with feature set Nav. + Facial). Bold indicates the highest performance.}
    \label{tab:results}
\end{table}